\documentclass[conference]{IEEEtran}
\ifCLASSINFOpdf
\else
\fi
\usepackage{adjustbox}

\usepackage{orcidlink}
\usepackage{amsmath}
\usepackage{amssymb}
\usepackage{algorithm}
\usepackage{algpseudocode}
\usepackage{booktabs} % For prettier tables
\usepackage{pifont}   % For check marks
\usepackage{amssymb}  % For empty checkbox
\usepackage{adjustbox}

\newcommand{\cmark}{\ding{51}} 
\newcommand{\xmark}{\ding{55}}  
\begin{document}
\title{Critic Loss for Image Classification}

\author{\IEEEauthorblockN{Brendan Hogan Rappazzo, Aaron Ferber, Carla Gomes}
\IEEEauthorblockA{Department of Computer Science\\
Cornell University\\
Ithaca, New York 14850\\
Email: \{bhr54, amf272, cpg5\}@cornell.edu}
% \author{\IEEEauthorblockN{Brendan Hogan Rappazzo}
% \IEEEauthorblockA{Department of Computer Science\\
% Cornell University\\
% Ithaca, New York 14850\\
% Email: bhr54@cornell.edu}
% \and
% \IEEEauthorblockN{Aaron Ferber}
% \IEEEauthorblockA{Department of Computer Science\\
% Cornell University\\
% Ithaca, New York 14850\\
% Email: amf272@cornell.edu}
% \and
% \IEEEauthorblockN{Carla Gomes}
% \IEEEauthorblockA{Department of Computer Science\\
% Cornell University\\
% Ithaca, New York 14850\\
% Email: gomes@cs.cornell.edu}
}


%Telephone: (800) 555--1212, Fax: (888) 555--1212}


\maketitle

\begin{abstract}
Modern neural network classifiers achieve remarkable performance across a variety of tasks; however, they frequently exhibit overconfidence in their predictions due to the cross-entropy loss. 
Inspired by this problem,  we propose the \textbf{Cr}i\textbf{t}ic Loss for Image \textbf{Cl}assification (CrtCl, pronounced Critical). 
CrtCl formulates image classification training in a generator-critic framework, with a base classifier acting as a generator, and a correctness critic imposing a loss on the classifier. The base classifier, acting as the generator, given images, generates the probability distribution over classes and intermediate embeddings. 
The critic model, given the image, intermediate embeddings, and output predictions of the base model, predicts the probability that the base model has produced the correct classification, which then can be back propagated as a self supervision signal. Notably, the critic does not use the label as input, meaning that the critic can train the base model on both labeled and unlabeled data in semi-supervised learning settings. CrtCl represents a learned loss method for accuracy, alleviating the negative side effects of using cross-entropy loss.
Additionally, CrtCl provides a powerful way to select data to be labeled in an active learning setting, by estimating the classification ability of the base model on unlabeled data.
We study the effectiveness of CrtCl in low-labeled data regimes, and in the context of active learning. In classification, we find that CrtCl, compared to recent baselines, increases classifier generalization and calibration with various amounts of labeled data. In active learning,  we show our method outperforms baselines in accuracy and calibration. We observe consistent results across three image classification datasets.
\end{abstract}


% \fi
\IEEEpeerreviewmaketitle

\section{Introduction}
\label{sec:intro}

% There have been big advances in archtuecture leading to big results
In recent years, significant advances in the architecture  of deep learning models have led to the development of powerful automated systems \cite{gpt_openai,alpha_go,alpha_fold}. 
% 
% Core efficacy is still the optimization of huge function
Central to the efficacy of these models is the rigorous optimization of millions of parameters with respect to a given loss function \cite{adamw_opt}.
While crafting clever network architectures and loss functions has proven important \cite{attention_need,focal_loss}, often, the best empirical results come from letting large models learn as end-to-end as possible for the task at hand \cite{end_to_end_cars,blip}. That is, where the model is trained to directly map inputs to outputs, and directly optimize for the relevant metric, learning the entire sequence of transformations required without relying on human-crafted intermediate steps or features. End-to-end training enables the model to autonomously discover the most effective representations and relationships for the task at hand.

\begin{figure*}[!t]
  \centering
   \begin{adjustbox}{width=.6\linewidth,center}
  \includegraphics{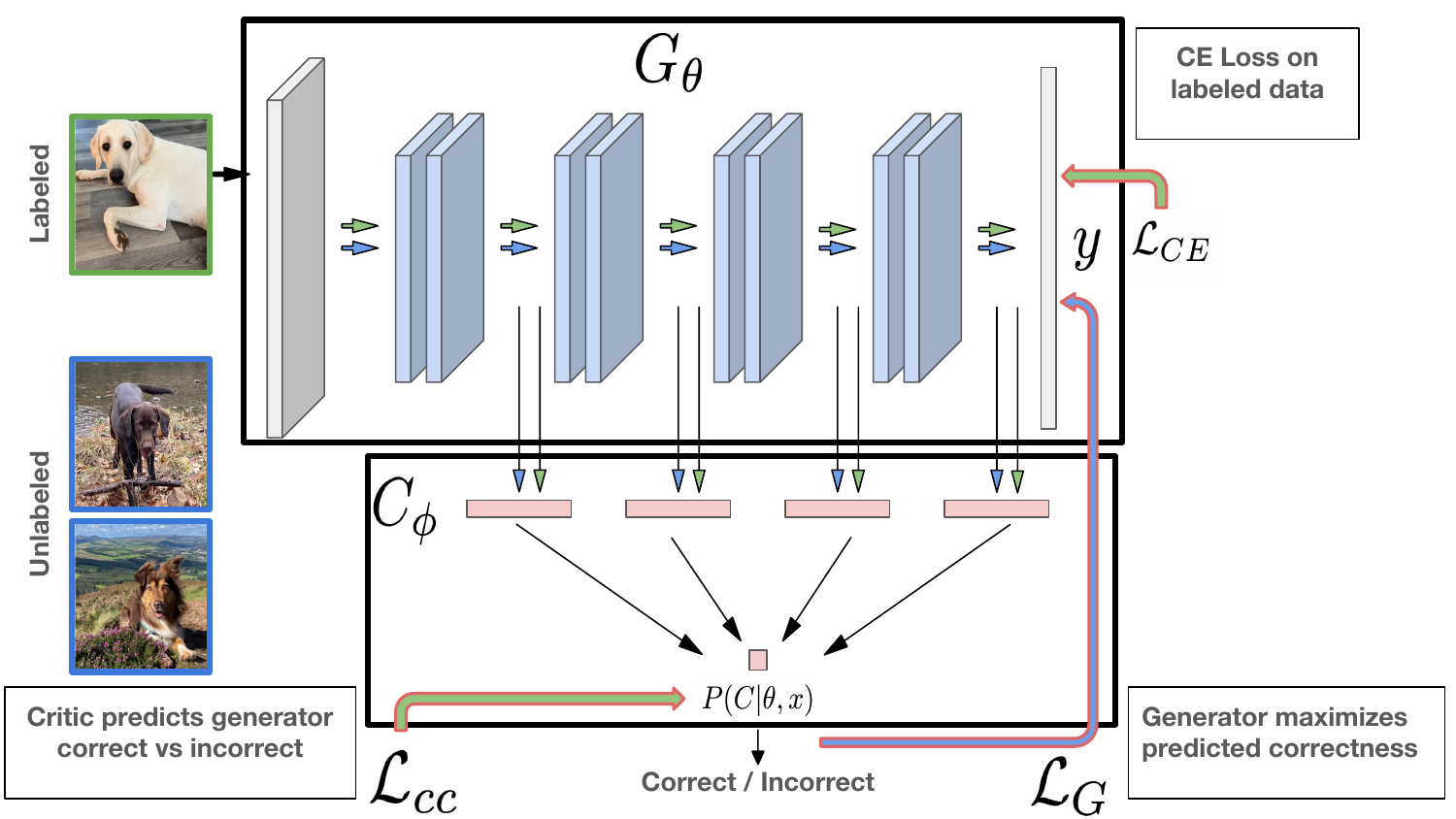}
  \end{adjustbox}
  \caption{A schematic of CrtCl, the classifier $G_{\theta}$ takes in images and produces intermediate representations and class probabilities, where the $\mathtt{ARGMAX}$ of the probabilities is the classification. The critic network, $C_{\phi}$, takes in the representations of $G_{\theta}$ and predicts whether $G_{\theta}$ classifies an example correctly. Once trained, the critic network can be used as a learned loss on both labeled and unlabeled data to train the generator to be more correct, while avoiding miscalibration from cross-entropy loss. Further, the critic model's prediction on unlabeled data points can be used to suggest misclassified points for active learning.}\label{fig:city}
\end{figure*}
% However - when what we truely care about isn't differentiabl must use proxies - for example accuracy 
In order to directly optimize for the most relevant metric for a given task, the metric itself should be represented as a differentiable loss function. For regressions tasks, this presents no problem, the metric of interest (Mean Absolute Error, Mean Squared Error etc.) is directly differentiable. However, for tasks like classification, the true metric of interest, accuracy, is not. To remedy this, proxy loss functions like cross-entropy, which encourages the output probability of the network to match a one-hot encoding of the ground truth label, are used. This works extraordinary well in practice, and large scale image classification networks have reached or surpassed human level performance on a host of tasks \cite{mao2023crossentropy}. However, this is not purely end-to-end learning, and while cross-entropy is a good proxy for accuracy, there are unintended side effects. 

% However - this isnt so end-to-end and has unintended side effects 

% Namely ill-calibration, and over fitting (less generalizable functions), whcih can have downstream effects of bad active learning 
One particularly well-studied side effect, with significant impact on the industrial use of neural networks, is calibration \cite{calibration_survery, end_to_end_cars}. Calibration is the notion or measure of how well the probability output of a network, matches its accuracy rate.  Modern large-scale networks tend to be ill-calibrated in that they are overconfident in their predictions - an artifact of the one-hot encoding of cross-entropy loss \cite{killian_calibration}. Calibration is extremely important in practice because it allows the users of a model to correctly understand the uncertainty in a prediction. Further, ill-calibration can affect down-stream tasks, in particular it can effect the efficacy of using a model's output in risk analysis, and active learning \cite{healthcare_calibration}. In active learning, a well calibrated model can be used to do uncertainty sampling by making predictions on unlabeled data, and it's most uncertain predictions can be used to guide which next samples to collect \cite{settles2012active}.

% Inspired by this we propose critic loss, which end to end learns to encourage the model to be "correct" what we truely care about, and learn the loss function. 
Inspired by this problem, we introduce a new learned loss function called Critic Loss For Image Classification (CrtCl, pronounced Critical). CrtCl formulates image classification as directly training a classifier to optimize accuracy by simultaneously training a critic to estimate accuracy. From one perspective, CrtCl treats image classification as a generator critic two-player game, where the base classification model \textit{generates} features and class probabilities, and the \textit{critic} learns to distinguish between correct or incorrect predictions. As the critic network is a differentiable neural network, it can be used as a loss function to teach the generator how to fix incorrect predictions to be correct.  As a result, CrtCl allows for end-to-end training of the accuracy metric, which leads to more generalizable and better-calibrated models.

% In implementation we find model still need standard cross-entropy to converge, but our loss can be a powerful regularizer, allowing the model to learn better features. 
We demonstrate the effectiveness of CrtCl for active learning. In many real-world settings, abundant data exist but few are labeled. Furthermore, labeling data can be expensive, especially in medical and cybersecurity domains \cite{expensive_medical,medical_li_hal-ia_2023, malware_expensive}. It is thus necessary to have models that can learn effectively with fewer labels, and also strategically identify which samples to label next. CrtCl, coupled with standard cross-entropy loss, outperforms baseline methods for active learning for image classification on three separate data sets, particularly in realistic low-labeled data regimes. We show that the critic model can be used for semi-supervised learning by applying the critic loss to unlabeled samples and for selecting the most informative points to label to improve accuracy. We also show that the models trained with our method tend to be better calibrated than other methods. Lastly, we run several ablation experiments to understand the effect of CrtCl's use as an auxiliary loss function, as a method to actively sample the next points to be labeled, and how these facets influence each other as a joint method.  We show that our method, even just as an auxiliary loss function, outperforms baseline methods in terms of accuracy and network calibration.

% % Our contributions
% Our contributions are as follows: 
% \begin{itemize}
%     \item We introduce our Critic Loss For Image Classification (CrtCl) method which aims to learn end-to-end the accuracy metric.
%     \item We show that our method can be used to guide active learning, and in a semi-supervised learning framework, and outperforms state-of-the-art techniques, allowing for both better learning in low-data regimes, and for better selection of the next best data to have labeled.
%     \item We show that our method consistently leads to better calibrated networks, leading to more 
% \end{itemize}

% Our contributions
Our contributions are 1) we introduce our Critic Loss For Image Classification (CrtCl) method which aims to \textbf{train classifiers to optimize accuracy end-to-end}. 2) We show that CrtCl can be deployed in \textbf{active learning and semi-supervised learning settings}, outperforming recent baseline techniques, and allowing for both better learning in low-data regimes, and better selection of the data label. 3) We show that CrtCl tends to produce \textbf{better calibrated classifications} meaning that end-users have a better grasp on the model's uncertainty.
    % \item We show that our method can be used to guide active learning, and in a semi-supervised learning framework, and outperforms state-of-the-art techniques, allowing for both better learning in low-data regimes, and for better selection of the next best data to have labeled.
    % \item We show that our method consistently leads to better calibrated networks, leading to more 
% \end{itemize}

\section{Related Work}
\subsubsection{Calibration}
Calibration measures the alignment between a model's uncertainty and observed probabilities, and it is often measured by the expected calibration error (ECE) \cite{ece}. It has been shown that modern neural networks, especially larger models with low classification error, tend to be ill-calibrated and are overconfident in their predictions \cite{network_calibration}. It has been show that calibration is particularly relevant in online settings \cite{online_calibration} and for structure predictions \cite{kuleshov_calibrated_2015}. 

%Interestingly, there seems to be a connection between different regularization types and calibration, for example Dropout \cite{dropout} has been shown to potentially impact calibration \cite{gal2016dropout}.

Several loss-based methods have been introduced to improve calibration such as a reguarlization term that also occasionally improves generalization \cite{penalize_overconfident}, a focal loss \cite{focal_cal}, a relaxation of ECE \cite{learn_ece}, a calibration-inducing kernel \cite{kernel_cal}, and label smoothing \cite{original_label_smoothing}. Label smoothing, which relaxes the hard one-hot loss of cross-entropy to use ``smoother'' labels, has been further shown to improve model calibration \cite{when_does_label_smoothing_help}.
Label smoothing and calibration also relate to work in knowledge distillation (KD) \cite{kd_distill_original}, in which the authors trained a secondary smaller network to predict the output logits of a larger network.  There have been many works on different version of KD, including self-distillation \cite{self_distill}. With the work in \cite{mixup_calibration}, the authors study the effect that image augmentation combined with loss functions, such as MixUp \cite{zhang2018mixup}, can improve calibration. 

% Further, \cite{original_label_smoothing} introduced the concept of label smoothing, which relaxes the hard one-hot loss of cross-entropy to "smoother" labels. 
% This was further studied in \cite{when_does_label_smoothing_help} which showed that smoothing labels can improve model calibration. 
% Label smoothing and calibration also relate to work in knowledge distillation (KD) \cite{kd_distill_original}, in which the authors trained a secondary smaller network to predict the output logits of a larger network. This training allows for much smaller networks to exhibit accuracy similar to that of the larger network, despite their smaller size. There have been many work on different version of KD, including self-distillation \cite{self_distill}. 

\begin{table}[t]
    \centering
    \resizebox{.5\textwidth}{!}{
    \begin{tabular}{l |c|c|c|c|c|c}
        \toprule
        % \textbf{Method} & \textbf{Auxiliary Networks} & \textbf{Hyperparameters} & \textbf{Steps Per Epoch} & \textbf{Active Learning} & \textbf{Semi-Supervised} & \textbf{Calibration} \\
        \textbf{Method} & \textbf{\shortstack{Auxiliary\\Networks}} & \textbf{\shortstack{Hyper-\\parameters}} & \textbf{\shortstack{Steps Per\\Epoch}} & \textbf{\shortstack{Active\\Learning}} & \textbf{\shortstack{Semi-\\Supervised}} & \textbf{\shortstack{Calib-\\ration}} \\
        \midrule
        Label Smoothing \cite{original_label_smoothing} & 0 & 2+ & 1 & \xmark & \xmark & \cmark\\
        Temperature Scaling \cite{network_calibration} & 0 & 2+ & 1 & \xmark & \xmark & \cmark \\
        Learning Loss \cite{learning_loss} & 1 & 2 & 1 & \cmark & \xmark & \xmark \\
        TOD \cite{huang2022temporal}& 2 & 3 & 1 & \cmark & \cmark & \xmark\\
        PT4AL \cite{pretrain} & 1 & 2 & 1 & \cmark & \xmark & \xmark \\
        \midrule
        CrtCl (ours) & 1 & 2 & 2 & \cmark & \cmark & \cmark \\
        \bottomrule
    \end{tabular}
    }
    \caption{Comparison of Active Learning and Calibration Methods}
    \label{tab:active_learning_methods}
\end{table}

\subsubsection{Active Learning}
Active learning is a rich and well studied field, with many branches \cite{4}.  Methods have been proposed which select examples based on the predicted class probabilities \cite{25,26}, the difference between the top $k$ predicted classes \cite{19}, and the entropy of the predicted probabilities \cite{47, 19}. 
% Additionally, query-by-committee methods \cite{49,34,18} use the agreement or disagreement between many models to estimate uncertainty.
% Uncertainty based models generally rely on the output of a single model, but another actively studied area of active learning is the query-by-committee methods 

% Data-centric approaches aim to estimate the data distribution, and then choose samples that represent the full distribution. Some methods formulate this as an optimization problem \cite{57, 9, 15}, or cluster the data and pick representative points that cover the feature space \cite{37}. Other methods use the location of the labeled data relative to the unlabeled samples to decide how to best propagate the knowledge that has already been learned \cite{5, 16, 32}. 
% While other methods use clustering to pick data to label that covers the feature space \cite{37}. 

A relatively new class of methods most similar to CrtCl aims to quantify the expected improvement or loss of the current model to use as proxy for data selection \cite{44,12,21}. 
% Some methods use Monte Carlo Dropout to expensively estimate uncertainty \cite{14}. 
In the state of the art Temporal Output Discrepancy (TOD) \cite{19}, the prediction disagreement on the unlabeled data acts as a proxy to estimate uncertainty, and thus select samples. A similar method uses a generative-adversarial model to predict which samples belong to the labeled or unlabeled data sets  \cite{val}. 

Also similar to our work is a method that first trains a network on the pretext task of predicting image rotation, which requires no labels \cite{pretrain}, and uses this loss to select samples.

Most similar to our work, Learning Loss for Active Learning (LL) \cite{learning_loss}, uses an auxiliary network to predict the loss value of the base network for a given data point, and then uses this prediction on unlabeled data to select data likely to have a high loss value. Further, this method has been extended with more mathematical analysis, particularly for regression tasks \cite{ll_pp}. Additionally, similar methods have been explored for segmentation methods, particular to detect adversarial attacks \cite{besnier2021triggeringfailuresoutofdistributiondetection}.

The motivation of our work, to find an alternative loss function for image classification is also similar to work in energy models \cite{jem}, however we argue these models have a similar issue of optimizing a proxy loss function. 

\subsubsection{Semi-supervised Learning}
% Semi-supervised learning is well studied area of research with methods using graph-based models \cite{57_t}, transductive models \cite{18_t}.

% Mean Teacher methods \cite{47_t}, and Temporal Ensembling \cite{28_t}. 
Semi-supervised learning is well studied area of research with
the work by \cite{49_t} provides an excellent survey of modern Semi-supervised learning methods.

\subsubsection{Overview}
We compare several salient approaches in \autoref{tab:active_learning_methods}. Notably, CrtCl is the only approach geared towards improving both active learning and semi-supervised learning while also aiming to improve network calibration and uncertainty estimation, something especially important in low-labeled data regimes. These works were picked as baselines because of their similarity to our method, and because of their relatively recent publication and performance on active learning leader boards (\url{paperswithcode.com)}. 

\section{Method}
% \subsection{Overview}
In this section, we formalize CrtCl, which aims to improve model generalization and active learning. The method involves the formulation of image classification in a generator-critic framework. Here, the classification network acts as a generator, aiming to generate correct predictions with respect to the critic network. The critic network aims to discriminate between correct and incorrect predictions. We first describe the setup of the two networks and CrtCl's training algorithm. 

\subsection{Problem Statement}
In semi-supervised learning, we are given sets of labeled examples, $\mathcal{D}_L$, unlabeled examples $\mathcal{D}_U$, and a test set $\mathcal{D}_{Test}$. We aim to train a classification neural network $G_{\theta}$ that maximizes predictive performance on $\mathcal{D}_{Test}$ having only trained on $\mathcal{D}_L$ and $\mathcal{D}_U$. Here predictive performance is measured both in terms of accuracy as well as model calibration. In the active learning setting, we can also iteratively select unlabeled samples to label $x_i \in \mathcal{D}_U$. Ideally, we want a method which has high predictive performance with various amounts of labeled data in both active learning and semi-supervised learning settings.
% In the active learning setting we also aim to develop a method that can pick samples $x_i \in \mathcal{D}_U$, that once labeled lead to the highest accuracy on the test set for the base model. In both settings, we also aim for our method to allow for the semi-supervised training by training on $\mathcal{D}_U$. 

\subsection{Classification Network}
The classification network, represented as a generator is a function $G_{\theta}(x)$, parameterized by $\theta$,  maps an input $x \in \mathcal{X}$ (e.g., images) to a probability distribution over the class labels. Specifically, for a given input $x$, the classification network produces a vector $z_i \gets G_{\theta}(x)$ where $z_i \in \mathbb{R}^K$ where $K$ is the number of classes. Each element of this vector represents the predicted probability of the corresponding class. The predicted label $\hat{y}$ is then determined as the class with the highest probability, i.e., $\hat{y} = \mathtt{ARGMAX}(z_i)$.
The architecture of $G_{\theta}$ can be varied; however, for this work we use convolutional neural networks for image classification with more implementation specifics provided in the experimental section. 

\subsection{Critic Network}
The Critic is a function $C_{\phi}(\cdot)$, parameterized by $\phi$, which operates on the feature set generated by the classification network. For a given input $x$, the classification network $G_{\theta}(x)$ also produces a feature set $F_{\theta}(x) \in \mathbb{R}^M$, where $M$ represents the dimensionality of the feature space. The Critic then evaluates these features, $C_{\phi}(F_{\theta}(x))$, and outputs a scalar value. This value quantifies the Critic's confidence that, the classification network's prediction, equals the ground truth prediction $y$, that is that $y = \mathtt{ARGMAX}(G_{\theta}(x))$. 

There are many options for how and which features $F_{\theta}$ are passed to $C_{\phi}$, which is largely dependant on the architecture of $G_{\theta}$. Following the learning loss module introduced in \cite{learning_loss}, our Critic network takes in several intermediate feature layers from $G_{\theta}$, which are then passed through a global average pooling layer, a fully connected layer, and finally concatenated to produce an embedding of the generator's predictions, from which a fully connected layer outputs a single scalar value, representing the probability $G_{\theta}(x)$ is correctly predicting for $x$.

\begin{algorithm}[!t]
\caption{Training Procedure for Image Classification with Critic Network}
\begin{algorithmic}[1]
\State \textbf{Input:} Labeled dataset $\mathcal{D} = \{(x_i, y_i)\}_{i=1}^N$, Unlabeled dataset $\mathcal{D}_u = \{x_j\}_{j=1}^M$
\State \textbf{Initialize:} Classifier $G_{\theta}$, Critic $C_{\phi}$
\State \textbf{Parameters:} Learning rate $\eta$, epochs $E$, loss weight $\gamma$, stopping epoch $E'$
\For{epoch $= 1, \ldots, E$}
    \For{each $(x_i, y_i) \in \mathcal{D}$}
    \State $F_{\theta}(x_i), z_i \leftarrow G_{\theta}(x_i)$
    \State $\mathcal{L}_{CE} \leftarrow -\sum_{i=1}^{N}\sum_{c=1}^{C} y_{i,c} \log(z_{i,c})$
    \State Partition: $C_R, I \leftarrow \{\mathtt{ARGMAX}(z_i) = y_i\}, \{\mathtt{ARGMAX}(z_i) \neq y_i\}$
    \State $\mathcal{L}_{cc} \leftarrow \sum_{x \in I} \log C_{\phi}(F_{\theta}(x)) - \sum_{x \in C_R} C_{\phi}(F_{\theta}(x))$
    \State $\theta \leftarrow \theta - \eta \nabla_{\theta} \mathcal{L}_{CE}$
    \State $\phi \leftarrow \phi - \eta \nabla_{\phi} \mathcal{L}_{CC}$
    \If{epoch $< E'$}
    \State $\mathcal{L}_{G} \leftarrow -\gamma \sum_{x \in \mathcal{D}_u} \log C_{\phi}(G_{\theta}(x))$
    \State $\theta \leftarrow \theta - \eta \nabla_{\theta} \mathcal{L}_{G}$
    \EndIf

    \EndFor
\EndFor

\State \textbf{return} Optimized $G_{\theta}$ and $C_{\phi}$
\end{algorithmic}
\end{algorithm}

\subsection{Critic Loss Procedure}
The training procedure for critic loss uses two loss functions and two additional steps in addition to the standard cross-entropy loss. Let $\mathcal{X}$ be the input space (e.g., images) and $\mathcal{Y}$ be the output space (e.g., class labels). The training dataset is denoted as $\mathcal{D} = \{(x_i, y_i)\}_{i=1}^N$  and the unlabeled dataset is denoted as $\mathcal{D}_u = \{(x_i)\}_{i=1}^M$ where $x_i \in \mathcal{X}$ and $y_i \in \mathcal{Y}$,

It works as follows: for a given training input $x$, we compute the feature set $F_{\theta}(x)$ and output probabilities $z$ by computing $G_{\theta}(x)$. The cross-entropy loss $\mathcal{L}_{CE}$ for the classification task is given by:
\begin{equation}
\mathcal{L}_{CE} = -\sum_{i=1}^{N}\sum_{c=1}^{C} y_{i,c} \log(z_{i,c})
\end{equation}
where $y_{i,c}$ is the ground truth label per sample $i$ and per class $c$, and $z_{i,c}$ is the per $i$ and $c$.
Subsequently, predictions are categorized as correct or incorrect based on whether $\hat{y}$ matches the true label $y$. Let $C_R$ denote the set of correct predictions and $I$ denote the set of incorrect predictions. Then, the Critic loss $\mathcal{L}_{cc}$ is calculated using the Wasserstein distance, also known as the Earth Mover's distance, which is defined for probability distributions $P_r$, the distribution of incorrect samples, and $P_g$, the distribution of correct samples as:

\begin{equation}
    W(P_r, P_g) = \inf_{\gamma \in \Pi(P_r, P_g)} \mathbb{E}_{(r,g) \sim \gamma} [\|r-g\|],
\end{equation}
where $\Pi(P_r, P_g)$ denotes the set of all joint distributions $\gamma(r,g)$ whose marginals are $P_r$ and $P_g$ respectively, with $r$ being the prediction of the critic for incorrect samples and $g$ being for correct. Which is formalized by applying the Kantorovich-Rubinstein duality as:

\begin{equation}
     \min_G \max_{C \in \mathcal{C'}} \mathbb{E}_{x \sim P_r} [C(G(x)] - \mathbb{E}_{x \sim P_g} [C(G(x))]
\end{equation}
The set $\mathcal{C'}$ represents the space of 1-Lipschitz functions, ensuring that the discriminator is constrained to be a 1-Lipschitz function.

To enforce the 1-Lipschitz condition on the discriminator, \cite{arjovsky2017wassersteingan} proposed clipping the weights of the discriminator to a compact space $[-c, c]$, where $c$ is a hyperparameter. This can be formally represented as $w \leftarrow \text{clip}(w, -c, c),$ for every weight $w$ in the discriminator. Weight clipping directly constrains the capacity of the discriminator, ensuring that the gradient norms are bounded, which is a necessary condition for the 1-Lipschitz continuity. 

Thus using the Earth Mover's distance for our critic loss, we have $\mathcal{L}_{cc}$ formalized as:
\begin{equation}
\mathcal{L}_{cc} = \sum_{x \in I}  C_{\phi}(F_{\theta}(x)) - \sum_{x \in C_R} C_{\phi}(F_{\theta}(x))
\end{equation}
Here, $C_{\phi}(F_{\theta}(x))$ represents the Critic's assessment of the classification network's feature set, aiming to maximize the features of correct predictions and minimize incorrect predictions, and with $C_R$ and $I$ being the sets of correctly and incorrectly labeled examples respectively. 
In this step, $\mathcal{L}_{cc}$ is backpropagated to $C_{\phi}$, and $\mathcal{L}_{CE}$ is backpropagated to $F_{\theta}$.
\begin{figure*}[!t]
  \centering
  \includegraphics[width=.8\linewidth]{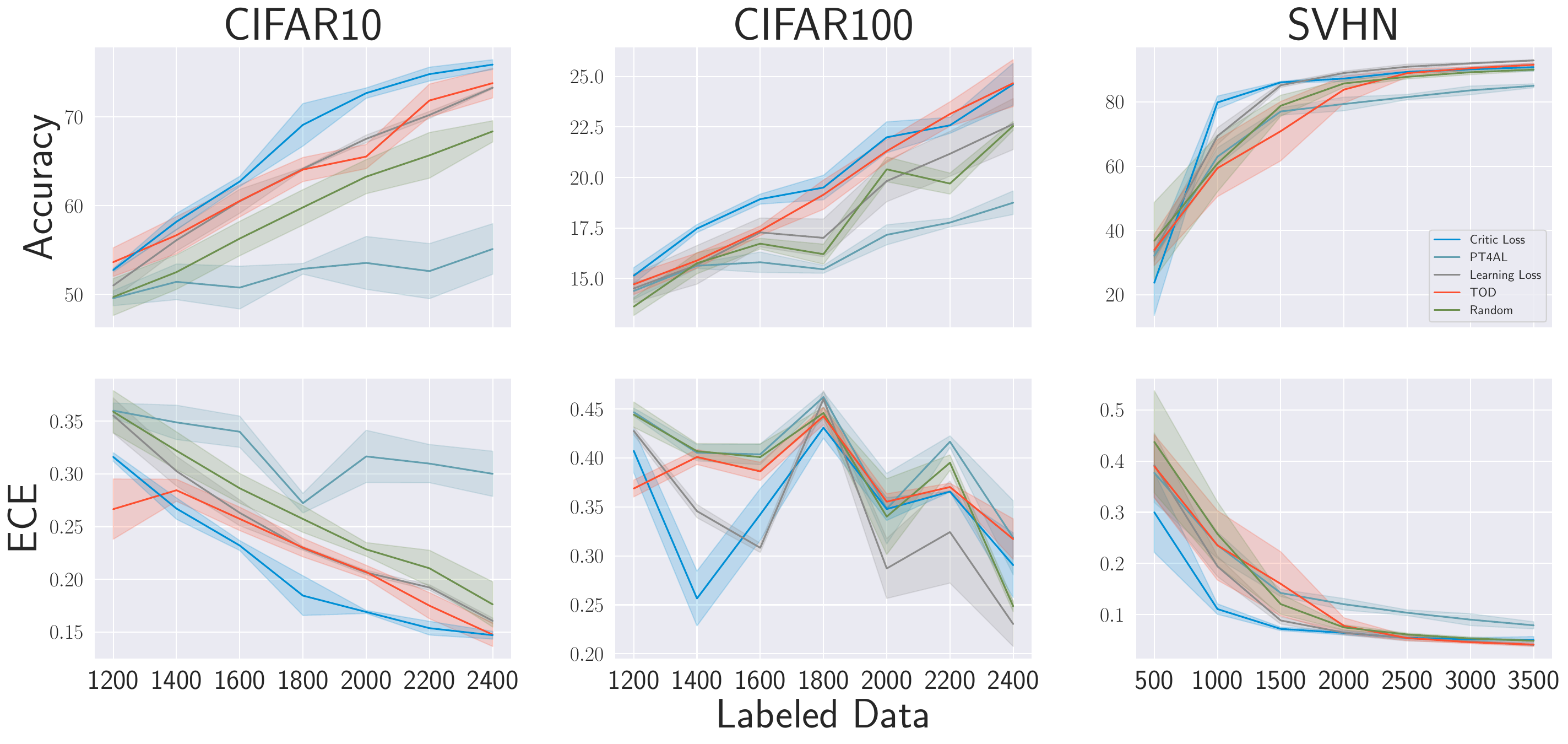}
  \caption{The accuracy and expected calibration error (ECE) results of our method, compared to Learning Loss \cite{ll_pp}, TOD \cite{huang2022temporal}, PT4AL \cite{pretrain}, and a standard training baseline for all three data sets. For all datasets, in the majority of active learning cycles, our method produced the more generalizable models (higher test accuracy), and better calibrated models (lower ECE).}\label{fig:exp_2_full}
\end{figure*}

In the semi supervised setting, we then sample a batch of data $\{x_1...x_b\} \in \mathcal{D}_u$. For each input $x \in \{x_1...x_b\}$, the generator's output $F_{\theta}(x)$,  is passed to the Critic. The Critic then assesses these outputs, and the following loss is computed
$\mathcal{L}_{G}$, is defined as:
\begin{equation}
\mathcal{L}_{G} = -\sum_{x \in \mathcal{D}_u} C_{\phi}(F_{\theta}(x))
\end{equation}
Here, $C_{\phi}(F_{\theta}(x))$ represents the Critic's assessment of unlabeled outputs from the classification network. By backpropagating this loss through the classification network, the model learns to adjust its predictions, aiming 
to correctly label these unlabeled data points. Note, in this work we didn't explore other types of GAN loss/training methods, as W-GAN's often lead to the most stable performance, however exploring other loss functions is a promising future work. 

\subsection{Active Learning}
In an active learning context, our model utilizes the Critic network to efficiently select samples from an unlabeled dataset for labeling. Let $\mathcal{D}_u$ denote the unlabeled dataset and $\mathcal{D}_L$ denote the labeled dataset. The active learning cycle proceeds as follows:

\begin{enumerate}
    \item For each unlabeled sample $x_u \in \mathcal{D}_u$, compute the feature set $F_{\theta}(x_u)$ using the classification network.
    \item Apply the Critic network to assess the probability the classifier is correct: $p_u = C_{\phi}(F_{\theta}(x_u))$.
    \item Rank the samples in $\mathcal{D}_u$ based on $p_u$, identifying those least likely to be correct.
    \item Select a subset $\mathcal{S} \subset \mathcal{U}$, comprising $n$ samples with the lowest $p_u$ values.
    \item Obtain labels for the samples in $\mathcal{S}$ from a human oracle, resulting in a set of newly labeled pairs $\{(x_s, y_s)\}_{s \in \mathcal{S}}$.
    \item Update the labeled dataset: $\mathcal{D}_L \leftarrow \mathcal{D}_L \cup \{(x_s, y_s)\}_{s \in \mathcal{S}}$.
    \item Retrain the model with the updated dataset $\mathcal{D}_L$.
\end{enumerate}

This approach allows for a focused expansion of the labeled dataset, prioritizing samples that are likely to provide the most informative feedback for model retraining. By iteratively applying this process, the model can effectively improve its performance with fewer labeled examples.

\section{Experiments}
In this work we focus on the efficacy of our model in the domain of active learning for image classification. We test our method in an active learning setting, where the model starts with few labeled data, but over the course of several rounds picks new data to be labeled. We gauge the efficacy of our method in both low-labeled data regimes, and in its ability to pick the most informative samples for labeling. Lastly, we provide two ablation studies to tease apart how much of the performance increase is coming from the samples being selected, versus better features being learned from the loss function.

\subsection{Datasets}
For all experiments we use the SVHN \cite{svhn}, CIFAR10 and CIFAR100 \cite{cifar_Krizhevsky2009LearningML}, data sets. CIFAR10 and CIFAR100 have an available labeled data set of 50,000 images and a test set of 10,000 images, with 10 or 100 classes respectively that represent common objects. SVHN has 60,000 training images, and 10,000 test images, for 10 classes of pictures of house number digits. For all data set, the images were normalized to have zero-mean and unit variance. All images were cropped to 32x32 pixels. For the training sets random horizontal flipping was used as an augmentation.

\section{Baselines}
We compare our method to recent state of the art methods, Learning Loss (LL) \cite{learning_loss}, Temporal Output Discrepancy (TOD) \cite{huang2021semi} and Using Self-Supervised Pretext Tasks for Active Learning (PT4AL) \cite{pretrain}. 

LL uses an auxiliary network, which takes the features produced by the base network,  to predict the loss value the base network will have on a given data point. In this way it can be used as an auxiliary loss, by backpropagating it's error through the features of the base network. Additionally, it uses its estimates of loss for unlabeled data to suggest which data will be best to have labeled by considering the data with the highest estimates of loss. 

The TOD method uses the output of the base model over different optimization steps to estimate the loss value of the model on specific data points. This difference between outputs at different optimization steps can be used to estimate loss for unlabeled data, and thus can be used as a selection criteria. Additionally, this difference of estimates on unlabeled data, and is used to train the base model in an unsupervised/semi-supervised manner. 

PT4AL is a recent work that has show state-of-the-art results for active learning and works by pre-training an auxiliary model on the pretext task of predicting image rotation. Then, for the unlabeled data pool, the rotation loss is used as a proxy to sample which data point to be next labeled. The method gives impressive results for active learning, however it does not have a way to train in a semi-supervised manner. 

Lastly, for all data sets and settings we compare to a standard Resnet18, trained with only cross-entropy loss, and with a random selection of the next points to include at each active learning cycle.

\subsection{Implementation Details}
Following \cite{huang2022temporal} for all experimental settings, we use a Resnet18 architecture. We use the last four feature layers of the network, as well as the input image and output probability distribution as the extracted features to train the critic network. The critic network takes these features as input, and use Global Average Pooling to pool them all. Then a linear layer is applied to each pooled feature layer, to project to a 128 dimensional space. Then each 128 dimensional space is concatenated, and a single linear layer projects the output to a single scalar value. The ReLU activation is used for all layers except the final layer.  

For $\mathcal{L}_G$, we find our method performs optimally, and only backpropagate to the last layer of the generator (which then is backpropagated to all other layers) instead of directly propagating to all feature layers. For the base network, in all cases we use Stochastic Gradient Descent as the optimizer with a learning rate of 0.1, and a momentum of 0.9. For the critic network we use the AdamW optimizer with a learning rate of 0.001. We train for 200 epochs, and decrease the learning rate by a multiplicative factor of 0.1 at epoch 160. For CIFAR10 and CIFAR100 we use a $\gamma$ value, which is the weighting of the $\mathcal{L}_G$, of 0.04, for SVHN we use 0.02. Additionally similarly to \cite{learning_loss}, we set $E'$,  the epoch we stop using $\mathcal{L}_G$, to 130 for all experiments.

For the active learning setting we start with a randomly initialized set of size $k$, where $k$ is 1200, 1200, and 500 for CIFAR10, CIFAR100, and SVHN respectively. We then train for 7 cycles, where for each cycle we add $t$ more labeled samples to the labeled training data set according the the active learning selection criteria, where $t$ is 200, 200, and 500 for CIFAR10, CIFAR100 and SVHN respectively. In this way, we aim to study the efficacy of our model in extremely low-labeled data regimes, and where each iteration very little labeled data is added to better mimic real-world settings. At the end of each cycle, the test accuracy and expected calibration error were recorded. All experiments are run for a total of 3 randomized trials, and the mean results with the +/- standard deviation are reported. 

For all other hyperparameters and baselines, we follow the methods described in \cite{learning_loss}, \cite{huang2022temporal} and \cite{pretrain}.

All experiments were run on 2xA100 NVIDIA GPUs, and while the time varied from experiment to experiment a single trial took approximately 1 hour of compute time.

\begin{figure}[!t]
  \centering
  \begin{adjustbox}{width=1.2\linewidth,center}
  \includegraphics{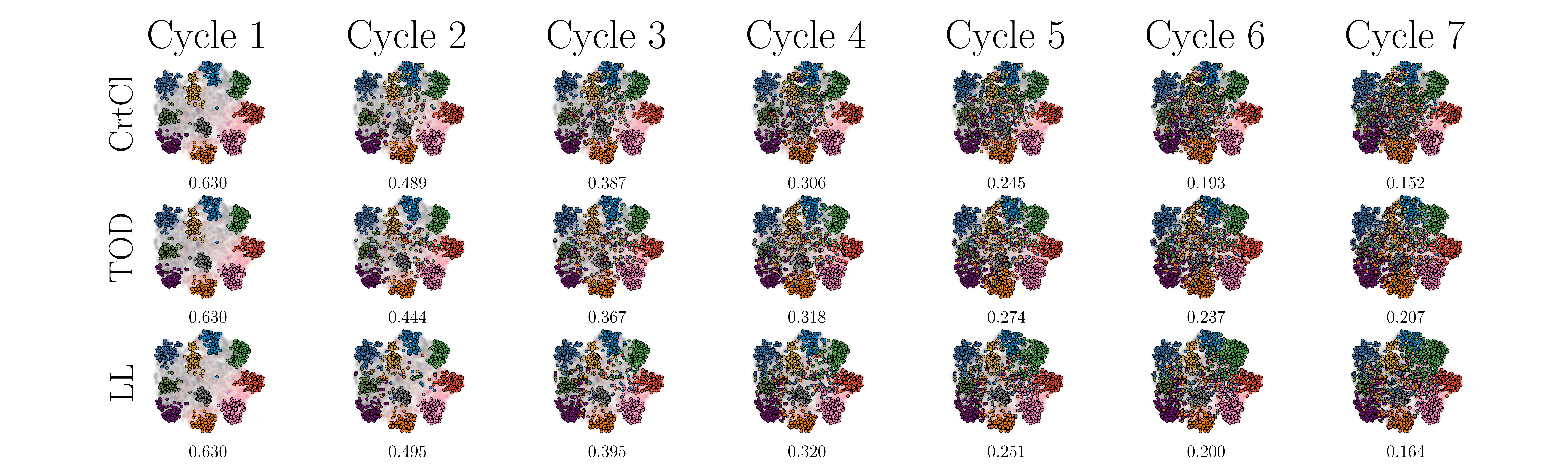}
  \end{adjustbox}
  \caption{For this setting we took a standard network trained with cross-entropy on CIFAR10, and produced t-SNE embeddings for the entire data set. We then show, for the highest performing methods, at each cycle which data points are selected by each method to be labeled. Fully opaque points represent those chosen to be labeled, and the unlabeled points are shown as highly transparent. Additionally we compute the mean silhouette score for each cluster. Intuitively, higher scores indicate the model is selecting points more similar to the data already in the labeled set. Whereas, the lower the score, the more the model is selecting more diverse samples that are spread out further in the feature space. We observe that early on in training, our model selects more similar samples, and later on in training it achieves the lowest clustering score, indicating it selects the most diverse samples per class.}\label{fig:exp3}
\end{figure}

\subsection{Evaluation}
The primary metric of interest is accuracy on the held out test set for each dataset to gauge model generalization. Further, we are also interested in model calibration which we measure using the expected calibration error (ECE). 
The expected calibration error (ECE) is a common metric for assessing the calibration of probabilistic models in classification tasks, where the lower ECE, the better the model is calibrated. Given a model with $c$ classes, we divide predictions into $M$ bins based on their predicted confidence. Let $B_m$ represent the set of indices of samples that fall into bin $m$, for $m = 1, \ldots, M$. The ECE is calculated as follows:

\begin{equation}
\text{ECE} = \sum_{m=1}^{M} \frac{|B_m|}{N} \left| \text{acc}(B_m) - \text{conf}(B_m) \right|
\end{equation}

\subsection{Active Learning Setting}
In \autoref{fig:exp_2_full} we compare the accuracy of CrtCl, the learning loss \cite{learning_loss}, TOD \cite{huang2022temporal}, PT4AL \cite{pretrain}, and a network trained with cross-entropy on randomly-selected active learning samples. As can be seen, for almost all cycles in all data sets our method outperforms the baselines in terms of accuracy, especially in the very low-labeled data regimes. We also show the resulting ECE for each method, which again shows in almost all cases that are model has the lowest calibration error. Showing that the learned classifier is better at estimating it's own uncertainty. 

\begin{figure}[!t]
  \centering
  \includegraphics[width=.95\linewidth]{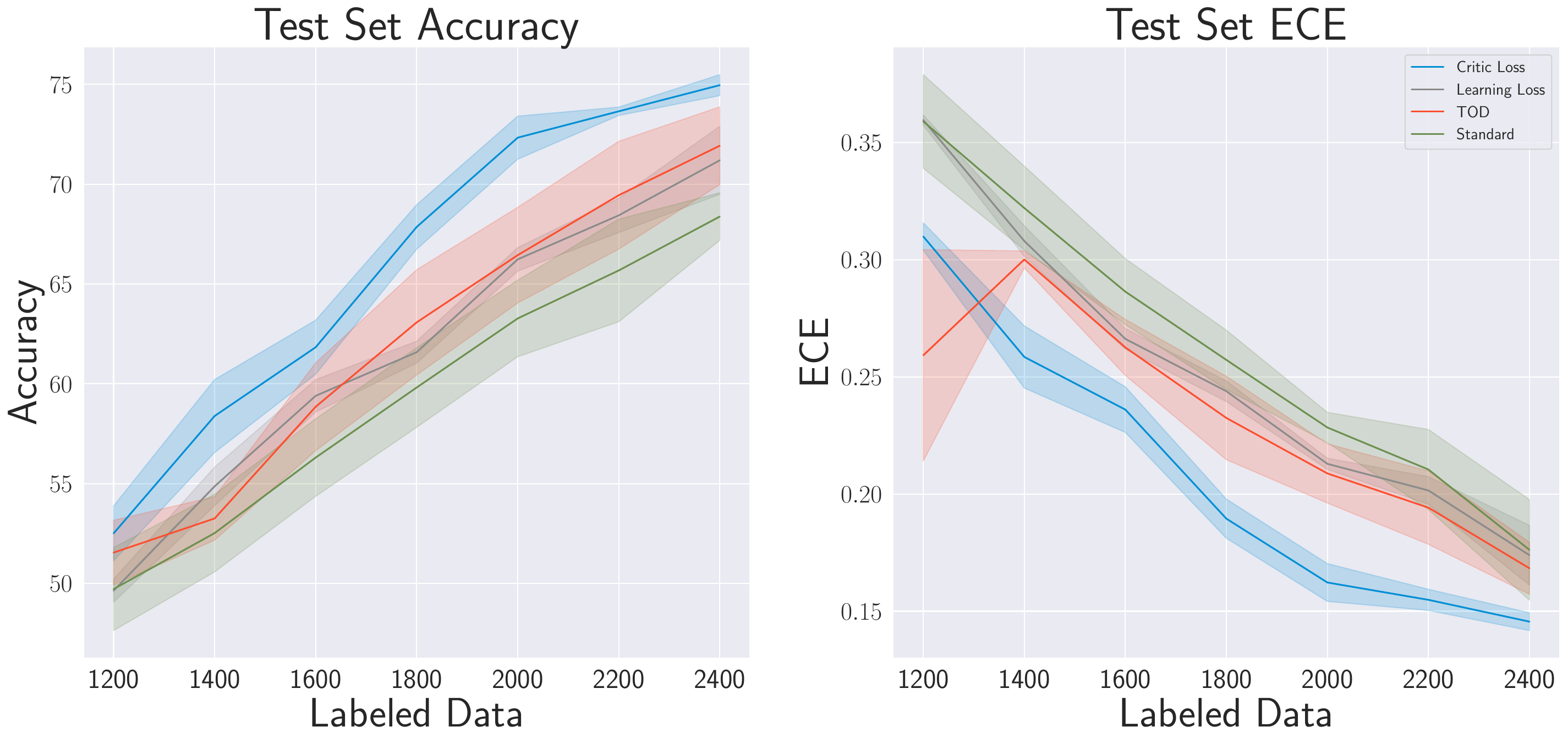}
  \caption{Our first ablation study which, for all methods, samples the next data points to be labeled randomly, but still uses the auxiliary loss function of all methods for the CIFAR10 data set. It is shown that our loss functions leads to better generalization (higher accuracy), and better calibration (lower ECE).
  }\label{fig:abl1}
\end{figure}

\subsection{Clustering Analysis}
In our second experimental setting we aim to show both qualitatively and quantitatively how our method selects data points from the overall data set. To this end, we study the CIFAR10 data set specifically. We take a Resnet18 network, trained in a standard setting with just cross-entropy loss until convergence on CIFAR10. We then use this network to extract the output features for the entire training set of CIFAR10, and compute a t-SNE projection. We then, for each method, for each cycle for a single trial,  plot the data selection as seen in Figure \autoref{fig:exp3}. Further, for each method and cycle we compute the silhouette score, for each label. 

% The silhouette score is a clustering metric used to compute the efficacy of a clustering method. Its value ranges from -1 to 1, where, for a single data point, a high value indicates that the data point is well matched to its own cluster and poorly matched to neighboring clusters. It is defined for a single data point has a value $s$ such that 
% $s = \frac{b - a}{\max(a, b)}
% $.
% Where $b$ is the mean distance from that point to the nearest cluster is it not part of, and $a$ is the mean distance from the objects in its own cluster. We use the true ground truth labels as the cluster label. 

In this case, higher scores mean the method is selecting active learning points that are most similar to those already in the labeled data set. Whereas lower scores means the clusters are more spread out over the embedding space, indicating the model is selecting data points for each class that are more diverse. 

Although largely qualitative, this analysis provides insight into the mechanism behind our method, and the comparable baselines.

We observe that for the first three cycles our method's silhouette score is in between the other two, suggesting our method, in early stages prefers data points that reinforce current knowledge, whereas after cycle four it has the lowest score, indicating it is successfully exploring more of the embedding space for each class compared to other models.

\begin{figure}[!t]
  \centering
  \includegraphics[width=.95\linewidth]{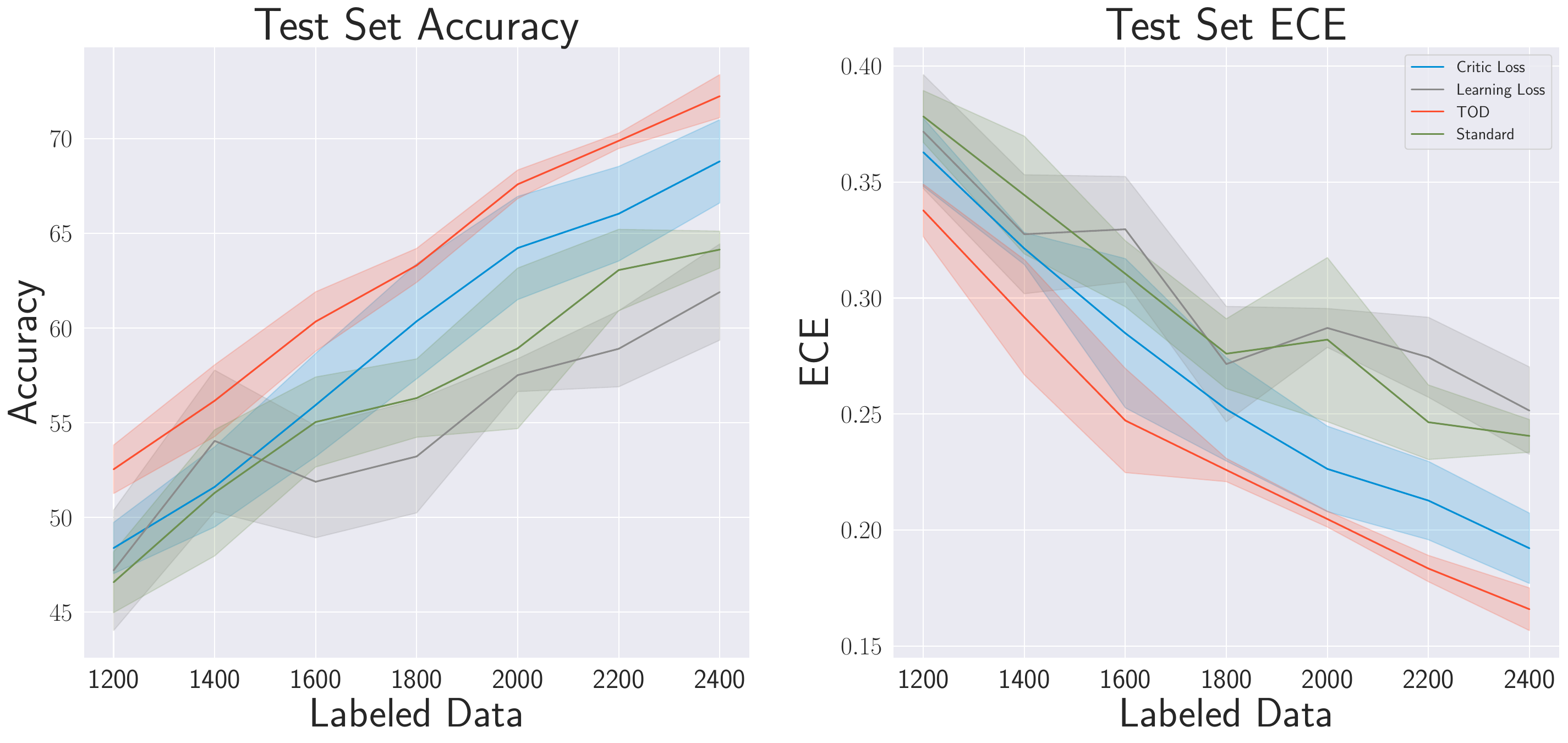}
  \caption{Our second ablation study which, for all methods, samples the next data points using the described methods, but does not backpropagate the auxiliary loss, for the CIFAR10 dataset. While our method still outperforms Learning Loss, TOD performs the best. Indicating that the benefits of our model are more entangled with the critic loss, which intuitively is expected given the nature of the generator-critic learning process. }\label{fig:abl2}
\end{figure}

\subsection{Ablation Studies}
All evaluated methods, besides PT4AL, have both an active learning selection mechanism, as well as a method to compute and backpropagate an auxiliary loss, whether on just the labeled set, or as a semi-supervised loss on the unlabeled data set. The natural question arises - how much of the improvement is from the auxiliary loss versus the data labeling selection mechanism. To this end, we designed and ran two ablation experiments where we test the efficacy of each model (besides PT4AL) with and without the auxiliary loss, and with and without the selection mechanism. 

\subsubsection{Auxiliary Loss Only }
In the first ablation study, we backpropagate the different auxiliary loss functions to the classification network, but when selecting points to label we randomly sample. Here, we aim to study how much of the benefit is purely coming from the CrtCl loss term. We plot the results for CIFAR10 in \autoref{fig:abl1} which shows improved test accuracy and calibration for all cycles. 

Further, the improvement is more exaggerated than when comparing to the full implementation. This suggests that our method, all else being equal, in just semi-supervised learning settings, may provide the best way to optimize a base network for highest accuracy, while preserving network calibration. That is, our generator-critic network loss function provides improvement over SOTA baselines, as well as standard cross-entropy loss in a standard image classification training setting.

\subsubsection{Active Learning Sampling Only}
In the second ablation study, we train the auxiliary networks as described for each method, but do not backpropagate the losses to the main network, and instead only use the selection mechanism for labeling data. In \autoref{fig:abl2}, we show the results for CIFAR10, which show that in this setting the TOD method outperforms all others, including our own. This result and the previous ablation study suggest that CrtCl performs well as an auxiliary loss and a joint loss and data selection method, whereas TOD performs better for data selection. 

%Intuitively, generator training is highly intertwined with the data selection in the generator-critic scenario, in that while the critic may select points that help improve the generator performance with respect to the critic, it is unable to select points that improve generator training with respect to cross-entropy loss.

\subsection{Trade-Off}
It should be noted that while our method yields superior results, it is more computationally expensive, requiring roughly double the number of gradient descent steps of standard deep learning training, as well as roughly twice the memory. While generally this range is acceptable, it is a significant trade-off and worth noting.

\section{Conclusion}
In this paper we introduce CrtCl, a novel learned loss method, which formulates training a classification network as a two player, generator-critic framework, where the base network generates features and probability distributions over classes, and the critic network produces a estimate that the generator network is correct. This critic network can be used to provide semi-supervision over unlabeled data, as well as to select data to be labeled in an active learning setting. Our method outperforms SOTA methods and standard baselines, in terms of accuracy, calibration and diverse sampling for three data sets.

\section{Acknowledgements}
This project is partially supported by the National Science Foundation (NSF); the Eric and Wendy Schmidt AI in Science Postdoctoral Fellowship, a program of Schmidt Sciences, LLC; the National Institute of Food and Agriculture (US-DA/NIFA);  the Air Force Office of Scientific Research) (AFOSR), and Toyota Research Institute (TRI).




\bibliographystyle{IEEEtran}


% \end{thebibliography}

\bibliography{main}

\end{document}